# Device-aware inference operations in SONOS non-volatile memory arrays

Christopher H. Bennett[1*], T. Patrick Xiao[1*], Ryan Dellana[1], Ben Feinberg[1], Sapan Agarwal[1], and Matthew J. Marinella[1]
1. Sandia National Laboratories, Albuquerque, NM 87185-1084, USA
{cbennet,txiao,mmarine}@sandia.gov
*These authors contributed equally

Vineet Agrawal[2], Venkatraman Prabhakar[2], Krishnaswamy Ramkumar[2], Long Hinh[2], Swatilekha Saha[2], Vijay Raghavan[3], Ramesh Chettuvetty[3]
2. Cypress Semiconductor Corporation, 198 Champion Court, San Jose, CA, 95134, USA
3. Cypress Semiconductor Corporation, Ste #120, 536 Chapel Hills Drive, Colorado Springs, CO, 80920, USA
ramesh.chettuvetty@cypress.com

*Abstract*— **Non-volatile memory arrays can deploy pre-trained neural network models for edge inference. However, these systems are affected by device-level noise and retention issues. Here, we examine damage caused by these effects, introduce a mitigation strategy, and demonstrate its use in fabricated array of SONOS (Silicon-Oxide-Nitride-Oxide-Silicon) devices. On MNIST, fashion-MNIST, and CIFAR-10 tasks, our approach increases resilience to synaptic noise and drift. We also show strong performance can be realized with ADCs of 5-8 bits precision.**

*Index Terms*-- SONOS, CTM, neural networks, edge inference

## I. Introduction

Non-volatile memory (NVM) arrays deliver low-latency and high-performance in-memory computing at projected throughputs of multiple tera-operations (TOPs) per second. They can also achieve femto-joule energy budgets per multiply-and-accumulate (MAC) operation by directly implementing within the analog circuit the vector-matrix multiplications (VMM) that are critical to neural network and scientific computing applications [1, 2]. While emerging analog NVM options such as filamentary/resistive RAM, phase-change memory (PCM) and conductive-bridge RAM (CB-RAM) are limited by relatively poor endurance, non-linearity, and cell-to-cell variation, foundry-accessible Silicon-Oxide-Nitride-Oxide-Silicon (SONOS) based charge-trap memory (CTM) cells offer an alternative pathway towards more standardized neural network realizations with 120× greater energy efficiency than an equivalent SRAM counterpart [3, 4]. Recently, similar three-terminal NOR Flash arrays implementing computer vision tasks were demonstrated [5], yet such systems are not at industry-standard complexity and do not consider several imperfect device issues. In this work, we interface a neural training library with device-aware inference modeling and propose how CTM devices can efficiently implement deep networks.

Prior work has considered the injection of noise during training to increase the resilience of the hardware to read noise, variability, and circuit non-idealities [6],[7]. We additionally investigate the impact of retention loss as measured in real devices and show that the same device-aware training methods can be applied to mitigate the effects of weight drift. In the following sections, we demonstrate that the following two approaches – in terms of software neural network and circuit architecture respectively – can mitigate device non-idealities:

- The use of noise regularization during training which implicitly counteracts generic noise disturbance, device-to-device variability, and charge decay phenomenon:
- The use of an appropriate 1T,1CTM cells in a dual array configuration to reflect analog weights, which counteracts device-to-device variability

At measured levels of charge decay and at small levels of intrinsic system noise, we demonstrate that SONOS arrays which reflect neural network weight matrices as analog physical quantities can yield inference accuracies that are relevant to modern machine-learning vision tasks.

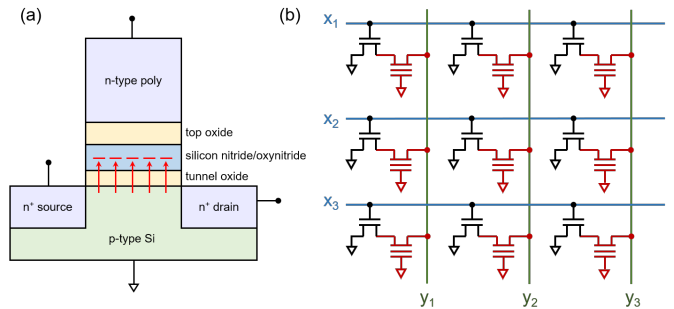

**Figure 1.** *(a) A three-terminal SONOS devices in which a synaptic weight is stored as a specified quantity of charge residing in a nitride charge trapping layer, (b) A computational array made up of SONOS devices for performing matrix-vector multiplication; inputs are applied as pulsed row voltages from the left and the outputs can be read out as currents at the bottom; a constant gate voltage is applied to all the SONOS devices (red). The 2T array uses an access transistor in every cell (black), used during programming..*

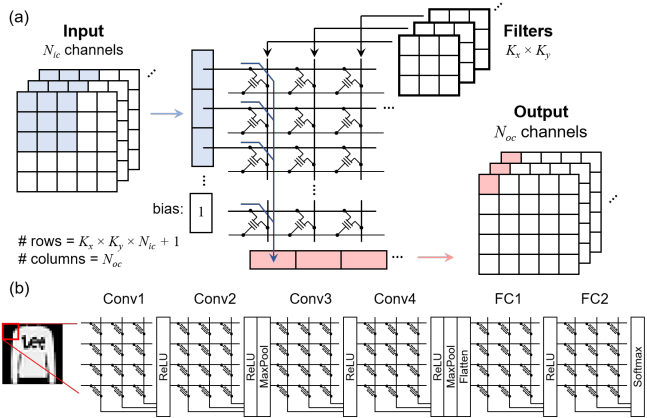

**Figure 2.** (a) *Sub-sets of an input (blue) are applied as vector inputs to an array containing a set of convolutional filters in a layer; each set (red) is then passed to a following array. (b) Our neural network, trained on MNIST or f-MNIST, where example images are convolved over a set of 4 layers followed by 2 fully connected (FC) layers.*

## II. SONOS DEVICE AND ARRAY

SONOS devices are highly suitable for analog memory with multiple levels because both program and erase operations can be achieved using Fowler-Nordheim (FN) tunneling, making it possible to target states with high precision in the drain current or threshold voltage. The ONO stack is a critical part of the SONOS memory cell, shown in Fig. 1(a). Charge is stored as electrons or holes residing in traps embedded in the nitride layer of the ONO stack. Depending on the type and amount of charge trapped, the threshold voltage ($V_t$) and current ($I_d$) of the cell can be placed at desired levels. The programming voltage and/or pulse width can be varied to set the device to different current levels. While the drain current can in theory be set by precise programming bias voltages, variation in these states is expected in practice, with a spread that depends on many device-level factors such as interface states in the SONOS stack, random dopant fluctuations and variations in charge trapping within the nitride. Moreover, the amount of variation can increase over time. So far, the realistic properties of the SONOS stack and their implications for a neural network inference accelerator have not been considered in detail.

We study three-terminal SONOS devices implemented in a dense array that holds a matrix of weight values, a small example of which is shown in Fig 1(b). The charge-trapping nitride layer can be programmed to a specified current level by applying a large voltage between the poly-cap gate and the drain. During inference, a fixed voltage $V_{G,read}$ is applied to the pass transistors of every cell in a bitwise fashion and the current drawn by each device is scaled by the input voltage, applied between the source and drain. The device currents along a column are summed by Kirchoff's current law and ultimately yield an output via the analog to digital converter (ADC). During programming [3, 4], our studied array uses the Control Gate (CG) having an ONO gate dielectric and a Select Gate (SG) having a SiO$_2$ gate dielectric. We also use two SONOS arrays to represent a matrix of real-valued weights, where each weight is represented by the difference of the currents drawn by a pair of SONOS cells. The currents drawn by corresponding columns in the two arrays are subtracted, and the result is again digitized using an analog-to-digital converter (ADC).

## III. CNN INFERENCE SIMULATIONS

Arrays of NVM devices can greatly accelerate matrix operations in neural networks, but with a potential loss in accuracy brought upon by non-ideal effects at the device level. Inference operations may be perturbed by cycle-to-cycle read noise, imprecise programming, device-to-device variation, and loss of state retention. In SONOS devices, these issues arise from imperfect control over the device dimensions, the spatial and energy distribution of desired and undesired traps, and the quantity of stored charge. Partly as a result, the implementation in NVM arrays of large convolutional neural networks (CNNs) – which generalize to hard real-time recognition tasks – has so far been limited [8]. Cognizant of this, we connect Keras, a library for training deep neural network models, and a new inference engine within the CrossSim platform, an open-source Sandia software that models crossbar arrays as highly parameterizable neural cores. In order to implement special bounding and clipping schemes during training, we have also used functions from the Whetstone library [9]. Our NVM-CNN uses an existing scheme to implement convolutional kernels on memory arrays, shown in Fig. 2 [10]. During inference, we import SONOS-specific parameters and inject cycle-to-cycle device noise to the synaptic weights (using CrossSim), while neuron noise is added during training (using Keras) onto the inputs to the rectified linear unit (ReLU) activation function. We bound the activations to a maximum value of 1.0, though we will also later consider a case with the bound removed. Implicitly, the use of training noise provides an effective form of neural network regularization and combats over-fitting [11]. The spread $\sigma_{neu}$ of the Gaussian noise introduced during training can be linked to the spread of cycle-to-cycle synapse noise $\sigma_{syn}$ during inference as:

$$(1) \qquad \sigma_{\text{neu}} = \sigma_{\text{syn}} (W_{\max} - W_{\min}) \sqrt{n}\, \gamma_{\text{act}}$$

where $\gamma_{act}$ is the average value of an activation calculated over $n$ synapses. When importing trained models, we select the device dynamic range used in each layer to contain the central 10–90% of the weight values in that layer; the extreme weight values are clipped to the maximum or minimum allowable device currents. By adaptively tuning the weight values that are represented by the device current levels, the effect of imprecision in the programmed currents (due to write errors, read noise, variability, or drift) can be minimized. In the results to follow, we find that clipping the extreme 20% of weight values during inference has a very small influence on the network's accuracy.

## IV. ACCURACY DEGRADATION DUE TO INTERNAL NOISE

As a first lens of analysis, we analyze the resilience of pre-trained neural networks – with and without noise regularization during training – to the disturbance of VMM operations by cycle-to-cycle read noise within the SONOS array. We assume two different noise models; one where Gaussian synaptic noise is dominated by an *additive* component (e.g. Johnson-Nyquist noise) that does not scale with the weight value, and one where the synaptic noise is dominated by a *proportional* component which scales with the device value (e.g. shot noise). For the present analysis, we do not implement a model of random telegraph noise (RTN), although this was done in [7]. Although additive and/or proportional Gaussian noise can cancel within layers, in deeper networks these effects compound and blur important convolutional features more dramatically than in shallower networks [12].

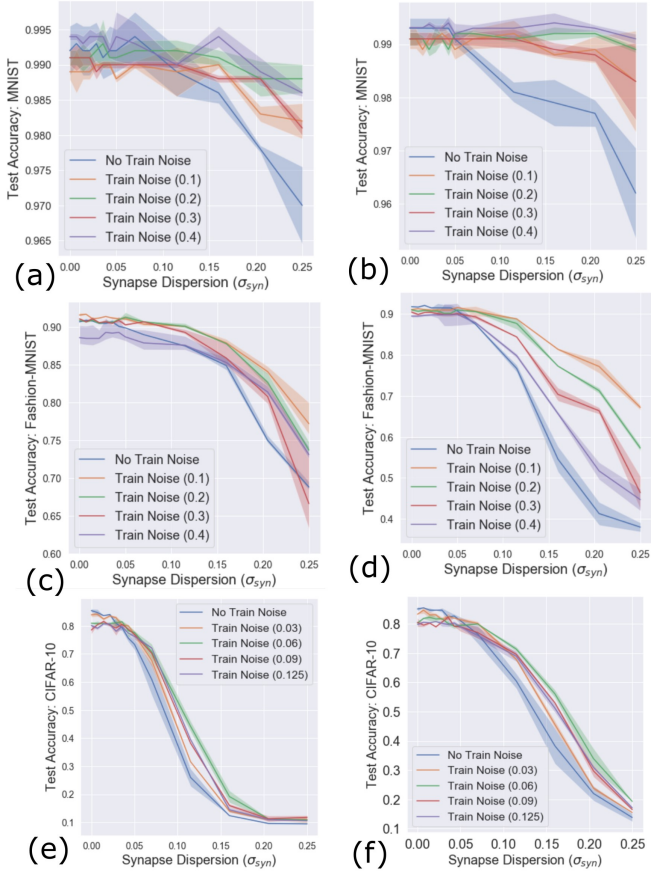

**Figure 3.** (a) *Performance on the MNIST task as additive read noise is injected for five neural network models trained with different levels neuron noise; for each, five different models were produced for statistical purposes with error bars as shaded regions; (b) MNIST task degradation as proportional read noise is injected; (c) fashion-MNIST degradation as additive noise is injected; (d) fashion-MNIST degradation as proportional noise is injected.; (e ) and (f) show additive and proportional for the CIFAR-10 task which is a set of 10 images from the ImageNet database with dimensions 32x32x3. In order to obtain all trained weights, batch size = 128, at task-specific learning rate was used; optimzer used was RMSProp for MNIST and Fashion-MNIST, and ADAM for CIFAR-10.*

As visible in Fig. 3(a) and (b), on the MNIST task, even a very large amount of read noise ($\sigma_{syn}$) fails to substantially degrade the network inference performance; robustness is nonetheless greater in the networks that were strengthened by noise regularized training. However, to a large extent the degree of resilience is a function of the task. With the same number of parameters (~500,000) between the models trained for the MNIST and fashion MNIST (f-MNIST) image recognition tasks, failure is more noticeable on the more difficult f-MNIST task, as shown in Fig. 3(c) and (d). The accuracy for neural networks trained both with and without noise regularization now falls substantially below the baseline without noise (~90%) on the fashion-MNIST task, decreasing below 85% when $\sigma_{syn} < 0.1$. Nonetheless, we find that noise-trained models at appropriate levels of $\sigma_{neu}$ can dramatically improve their resilience as compared to the model without any preparation for noise during training (*e.g.* blue curve). The best performing regularization approach does seem to vary based on the injected noise type; $\sigma_{neu} = 0.1$ (orange) has the greatest resilience to proportional noise, while $\sigma_{neu} = 0.4$ (purple) does best against additive noise.

Lastly, in Figure 3(e) and (f), we have analyzed performance on the CIFAR-10 task [13], which is a standard image recognition task in machine learning workflows and widely considered to be more difficult than either MNIST or fashion-MNIST. Notably, the images now contain 3 color channels and are of larger dimensionality (32x32 pixels). In the CIFAR-10 case, we see that noise degradation is even more severe, with $\sigma_{syn} < 0.05$ now necessary for good (>80%) performance on the task. In addition, at this harder task the noise regularization procedure has become more difficult and required smaller levels of $\sigma_{neu}$ in order to ensure convergence during the training stage. However, as visible, this noise regularization remains effective, with both proportional/additive noise cases showing regularized networks degrading more gracefully than the non-prepared example (blue). Although this task is easier than state-of-the-art tasks such as ImageNet, our analysis converges with a parallel analysis of internal noise impact on edge inference [14]. In addition, as in [15], we have found that proper weight scaling between neural network trained models and conductance mapped values is critical to resilience.

## V. ACCURACY DEGRADATION DUE TO CHARGE DECAY

In addition to synaptic read noise, loss of synaptic state retention due to the decay or leakage of charge from the trapping layer is also a potential source of neural network accuracy degradation during inference. A number of charge loss mechanisms have been identified in SONOS devices deployed for storage applications [16]. Among the main proposed mechanisms is a two-step leakage process that begins with Frenkel-Poole thermionic emission of trapped charge from the nitride layer, followed by tunneling through the oxide layer, with the assistance of oxide traps [17]. An alternative mechanism for threshold voltage drift, proposed in [18], is based on the lateral migration of charge within the nitride layer by means of thermal hopping from one trap to the next. The trapped electrons and holes migrate with different effective mobilities and recombine; over time, this process leads to a spatial charge distribution that causes a drift in the threshold voltage and subsequently in the drain current. The rate of current drift by both mechanisms depends on the number of program-erase cycles endured by the device, either through increasing the density of oxide traps or by increase the difference in electron and hole populations.

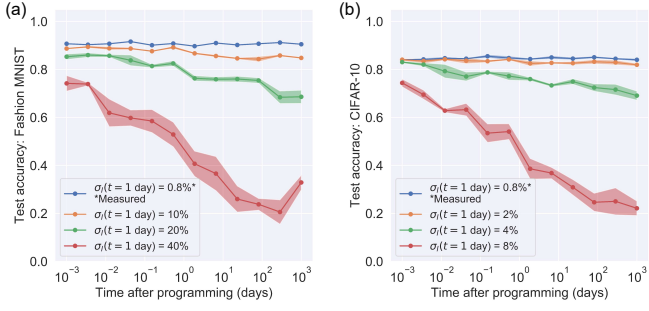

**Figure 4.** *The evolution of inference accuracy over time on (a) fashion MNIST and (b) CIFAR-10 as a result of charge decay. Using the measured SONOS device retention properties (blue), the degradation on both tasks is negligibly small. The performance curves corresponding to hypothetical devices of inferior quality, quantified here as the variability (as a percentage of the device dynamic range) in the programmed currents after one day, are also shown.*

We study the effect of charge decay on neural network inference by characterizing a prototype CTM array. SONOS devices were fabricated using 40nm SONOS eNVM technology by Cypress Semiconductor. The ONO stack layers are engineered to have the highly robust retention properties needed for analog memory. Notably, good retention can be achieved by trapping charge in deep traps rather than in shallow traps within the nitride layer. The fabricated devices were programmed to specified levels of drain current, and the retention properties as well as the standard deviation of the drain current over a group of devices were characterized over time. Measurements on the SONOS devices, conducted over a 24-hour period after programming, show that the drain current in all of the cells drift by an equivalent amount independent of the initially programmed current. Beyond this timeframe, the change in device properties was very small. Our scheme for representing positive and negative weights, which involves the subtraction of the current from two CTM arrays, naturally compensates for this type of uniform drift that does not depend on the initial current. Drift in the positive device, which increases the weight, is always canceled by the same drift in the negative device, which decreases the weight. On the other hand, since the charge loss mechanisms are stochastic, variable rates of current drift in different devices result in an increased spread or uncertainty in the device currents over time, whose effect is similar to an increase in the level of read noise, but does not vary from cycle to cycle.

In this work, we extrapolate the time dependence of the drain current $I(t)$ and its uncertainty $\sigma_I(t)$ by fitting the measured data to a stretched exponential function:

$$(2) \quad I(t) = I_0 + (I_\infty - I_0)\left(1 - \exp\left[-\left(\frac{t}{\tau}\right)^{T/T_0}\right]\right)$$

$$(3) \quad \sigma_I(t) = \sigma_{I,0} + (\sigma_{I,\infty} - \sigma_{I,0})\left(1 - \exp\left[-\left(\frac{t}{\tau}\right)^{T/T_0}\right]\right)$$

where $t$ is the time after the last programming event, $T$ is the operating temperature, $I_0$ is the initially programmed current with uncertainty $\sigma_{I,0}$, and $I_\infty$ is the saturated current at $t = \infty$ with a saturated uncertainty $\sigma_{I,\infty}$. This function has previously been used to model the time dependence of the threshold voltage $V_{th}$ of SONOS devices [18, 19]. In these works, $V_{th}$ was derived from the measured drain current using a fixed value of transconductance $g_m$, and the evolution of the threshold voltage spread was found to follow directly from the drift in the center value. To fit the measured retention properties in our SONOS devices, we use a characteristic decay time of $\tau = 1$ day. We also use a value of $T_0 = 2500$K from [19]. The decay time $\tau$ is physically derived from an activation energy as $\exp(-E_T/kT)$, where $k$ is the Boltzmann constant and $E_T$ depends on the spatial and energy distribution of traps, the composition of the ONO stack, and the number of program-erase cycles endured by the device.

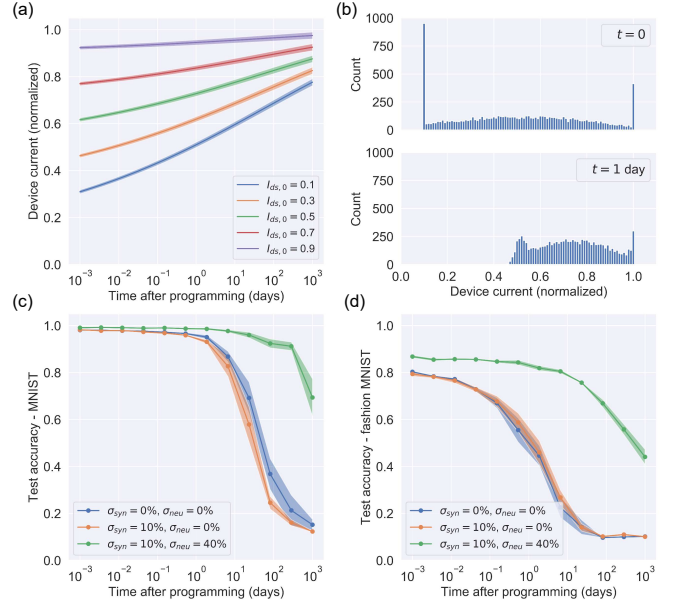

**Figure 5.** *(a) Modeled non-uniform drift characteristic of the device currents for previously published SONOS devices for several initial programmed values $I_0$. The spread $\sigma_I$ in weight values also increases with time. (b) The initial ($t = 0$) distribution of current values and the distribution after $t = 1$ day for a SONOS array trained for the MNIST task. The distribution is shown for the positive array of the fourth convolutional layer. The large peaks at the minimum (0.1) and maximum (1.0) values in the initial distribution arise because weights lying outside the center 80% range are clipped to these values. The evolution of the test set accuracy as a function of the time after programming are shown for (c) MNIST and (d) fashion MNIST. The injection of neuron noise during training ($\sigma_{neu} > 0$) increases resilience to synaptic read noise as well as the effects of weight drift.*

The measured SONOS devices show an initial variability in the programmed current $\sigma_I$ that is equivalent to 0.4% of the dynamic range, which increases to 0.8% after 24 hours. The blue curves in Fig. 4(a) and (b) show that this degree of variability, accompanied by a decay in the device dynamic range, does not visibly degrade the inference accuracy on the fashion MNIST task or the CIFAR-10 task. The result suggests that from the perspective of device retention, state-of-the-art CTM arrays optimized for analog memory are sufficient for long-term deployment in inference applications at the studied level of neural network complexity. The lack of any noticeable accuracy loss further suggests that the retention properties of these devices may also be sufficient for more difficult inference tasks than studied here that use larger, deeper, and more complex neural network topologies.

To investigate the implications of a more severe amount of charge decay using less optimized CTM devices, and to understand how drift effects might scale to more complex neural networks, we consider the hypothetical cases where the observed amount of charge decay has been amplified. These cases are shown by the orange, green, and red curves in Fig. 4: the same initial error of 0.4% of the dynamic range was assumed, but the uncertainty after one day was increased to several times the observed amount. At these inferior levels of device retention, the accuracy degradation resulting from drift and variability is more significant. Although the directional effect of a uniform current drift is fully canceled out as described above, it nonetheless reduces the useable dynamic range in each device, exacerbating the effect of both read noise and drift variability over time. The more challenging CIFAR-10 task is more sensitive to the effects of charge decay on the network weights, which follows from our finding that more challenging tasks are also more sensitive to read noise.

To examine the combined effects of non-uniform current drift and increasing current variability, we model the properties of a previously published SONOS device that was fit using the stretched exponential model in Equation (2). Here, we use a thermally activated decay constant $\tau = \tau_0 \exp(E_T/kT)$, using $\tau_0 = 8.0 \times 10^{-12}$ hours, $E_T = 0.85$ eV, and $T_0 = 2500$K, corresponding to a SONOS device measured in [19] that has endured 1000 program-erase cycles prior to deployment for inference. Unlike the uniform decay observed in our fabricated SONOS devices, we assume that all devices approach the same value of $I_\infty$ (set to the upper limit of the dynamic range) with the same time constant; thus, for the same time interval, devices with $I_0$ further away from $I_\infty$ will decay by a larger amount. Fig. 5(a) shows the current drift characteristics modeled by Equations (2) and (3) for several values of the initial current, normalized by $I_\infty$. The current drift that occurs on short time scales ($t < 1$ minute) is not explicitly shown. Fig. 5(b) shows how charge decay modifies the distribution of the weight values in the positive array in one of the layers of the neural network, trained for the MNIST task. After one day has elapsed after programming, the current values shift asymmetrically toward larger values. The use of two differential devices to represent a weight partially mitigates the effect of drift, but the cancellation is incomplete, since the amount of drift depends on the present value of the device current.

Fig. 5(c) and (d) show how this model of current drift in SONOS devices affects the test accuracy on the MNIST and fashion MNIST tasks, respectively. In the noise unprepared models, significant accuracy degradation begins after about one day for MNIST and within hours for fashion MNIST. Note that in this model, the variability $\sigma_I$ after one day – see Fig. 5(a) – is a considerably smaller fraction of the total dynamic range than the hypothetical cases shown in Fig. 4, suggesting that non-uniform decay and dynamic range compression can potentially play a large role in the performance degradation. Cycle-to-cycle read noise ($\sigma_{syn}$ = 0.1) further degrades performance, though the added effect is slight. Remarkably, the regularization effect provided by neuron noise injection during training ($\sigma_{neu} = 0.4$) considerably improves the system's resilience to drift; for both tasks, the point at which the accuracy drops to 70% is enhanced by multiple orders of magnitude.

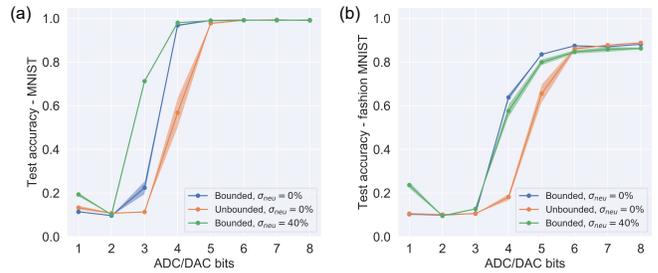

**Figure 6.** Test accuracy vs. ADC and DAC resolution for the (a) MNIST and (b) fashion MNIST tasks, with a model that is trained with a bounded (with and without training noise) vs. unbounded ReLU activation function.

## VI. Effects of Quantization

The conversion from the analog output of a computational memory array to a digital signal that can be routed to the next stage – and vice versa – is a costly operation in analog neuromorphic accelerators. While analog-to-digital converters (ADCs) and digital-to-analog converters (DACs) with high bit resolution are desired to maintain high precision in the neuron activation values, this typically comes at a large overhead in area, energy, and latency to the system, offsetting the critical advantages of computation in non-volatile memory arrays [1]. To this end, it is critical to determine the least amount of ADC and DAC precision that can still yield high neural network accuracy, and to design the system so that lower-precision, and thus lower-energy conversions can be tolerated.

Fig. 6(a) and (b) show the accuracy of the neural network as a function of the bit precision of the ADC and DAC, for the MNIST and fashion MNIST tasks, respectively. The blue curve shows the performance for CNNs that use a bounded ReLU function, which we have assumed for most of this paper, while the orange curve shows a standard unbounded ReLU function. The bounded ReLU is described by $f(x) = \min(\max(0, x), 1)$, and has an output in the range $(0, 1)$. Consequently, for all but the final softmax layer, the ADC and DAC range for the bounded system is set to $(0, 1)$, as any information outside of this range is discarded by the activation function. The ADC range for the final layer is calibrated by inspecting the output values from the relevant arrays. The range is chosen so that very large activation values are clipped, which slightly reduces the maximum attainable accuracy but reduces the precision requirement on the final layer ADC. For the unbounded neural network models, this calibration is performed on the output of every layer to set the range on the corresponding ADCs and DACs.

For both image recognition tasks, the use of a bounded ReLU function reduces the required ADC/DAC precision by one bit: down to four bits for MNIST and five bits for fashion MNIST. Any input to the bounded ReLU exceeding a value of 1 simply collapses to an output of 1; the ADC requirement is relaxed by eliminating the need to quantize larger array outputs. We also find that by constraining the activation function to operate within a finite range, the activation values tend to become more concentrated near the extreme values of 0 and 1. A one-bit reduction in the ADC precision potentially halves the energy budget. The known bounds on the activations also simplify the hardware design, as the ADC/DAC ranges no longer need to be data-dependent.

Further reduction in precision may be possible by training the network to more closely approach the limit of a binary distribution of activation values, which would imply that only one bit of ADC precision is needed. Noise injection during training, when combined with the bounded activation, improves the resilience to quantization effects for the MNIST task, but this is not seen for fashion MNIST.

VII. DISCUSSION

Our results so far imply that co-design between devices and inference accelerator can yield promising performance on state-of-the-art tasks used in the machine learning community. However, while our decay model so far used realistic SONOS data, our modeled noise is generic. In an extension of this work, we plan to incorporate device-specific parameters for these and combine them with realistic decay to in order to analyze whether equivalent-accuracy to software performance with emerging SONOS NVM is possible. In addition, an alternative approach to analog inference is the binary approach, in which weights and activations can be simplified to logic gates [20]. Prototypes of this concept with emerging filamentary and magnetic NVM devices have been proposed [21, 22]. Thus in the future we plan to rigorously benchmark analog inference versus competitor binary systems.

VIII. CONCLUSIONS

Neural network inference using SONOS-based in-memory-computation systems is fast, massively parallel, and highly energy-efficient compared to digital implementations, but its accuracy may be degraded by weight values that are imprecise or decay over time. We have shown that these architectures can retain high accuracy at conceivable levels of system noise by deploying device-aware training methods. We have also found that the retention properties of state-of-the-art SONOS devices are sufficiently ideal to support inference engines implementing large convolutional neural networks, without the need for frequent device updates.

ACKNOWLEDGEMENT

Sandia National Laboratories is a multimission laboratory managed and operated by NTESS, LLC, a wholly owned subsidiary of Honeywell International Inc., for the U.S. Department of Energy's National Nuclear Security Administration under contract DE-NA0003525. This paper describes objective technical results and analysis. Any subjective views or opinions that might be expressed in the paper do not necessarily represent the views of the U.S. Department of Energy or the United States Government.

REFERENCES

[1] Marinella M J, Agarwal S, Hsia A, Richter I, Jacobs-Gedrim R, Niroula J, Plimpton S J, Ipek E and James C D 2018 Multiscale co-design analysis of energy, latency, area, and accuracy of a ReRAM analog neural training accelerator *IEEE Journal on Emerging and Selected Topics in Circuits and Systems* **8** 86-101

[2] Burr G W, Shelby R M, Sebastian A, Kim S, Kim S, Sidler S, Virwani K, Ishii M, Narayanan P and Fumarola A 2017 Neuromorphic computing using non-volatile memory *Advances in Physics: X* **2** 89-124

[3] Agarwal S, Garland D, Niroula J, Jacobs-Gedrim R B, Hsia A, Van Heukelom M S, Fuller E, Draper B and Marinella M J 2019 Using Floating-Gate Memory to Train Ideal Accuracy Neural Networks *IEEE Journal on Exploratory Solid-State Computational Devices and Circuits* **5** 52-7

[4] Marinella M J, Jacobs-Gedrim R B, Bennett C, Hsia A H, Haghart D R, James C D, Vizkelethy G, Bielejec E, Agarwal S and Fuller E J 2019 Energy efficient neuromorphic algorithm training with analog memory arrays. Sandia National Laboratories Albuquerque United States)

[5] Merrikh-Bayat F, Guo X, Klachko M, Prezioso M, Likharev K K and Strukov D B 2017 High-performance mixed-signal neurocomputing with nanoscale floating-gate memory cell arrays *IEEE transactions on neural networks and learning systems* **29** 4782-90

[6] Klachko M, Mahmoodi M R and Strukov D B 2019 Improving Noise Tolerance of Mixed-Signal Neural Networks *arXiv preprint arXiv:1904.01705*

[7] He Z, Lin J, Ewetz R, Yuan J-S and Fan D 2019 Noise Injection Adaption: End-to-End ReRAM Crossbar Non-ideal Effect Adaption for Neural Network Mapping: ACM) p 57

[8] Xu X, Ding Y, Hu S X, Niemier M, Cong J, Hu Y and Shi Y 2018 Scaling for edge inference of deep neural networks *Nature Electronics* **1** 216

[9] Severa W, Vineyard C M, Dellana R, Verzi S J and Aimone J B 2019 Training deep neural networks for binary communication with the whetstone method *Nature Machine Intelligence* **1** 86-94

[10] Shafiee A, Nag A, Muralimanohar N, Balasubramonian R, Strachan J P, Hu M, Williams R S and Srikumar V 2016 ISAAC: A convolutional neural network accelerator with in-situ analog arithmetic in crossbars *ACM SIGARCH Computer Architecture News* **44** 14-26

[11] Noh H, You T, Mun J and Han B 2017 Regularizing deep neural networks by noise: Its interpretation and optimization p 5109-18

[12] Agarwal S, Plimpton S J, Hughart D R, Hsia A H, Richter I, Cox J A, James C D and Marinella M J 2016 Resistive memory device requirements for a neural algorithm accelerator: IEEE) p 929-38

[13] Krizhevsky A and Hinton G 2009 Learning multiple layers of features from tiny images

[14] Yang T-J and Sze V 2019 Design Considerations for Efficient Deep Neural Networks on Processing-in-Memory Accelerators *arXiv preprint arXiv:1912.12167*

[15] Rasch M J, Gokmen T and Haensch W 2019 Training large-scale ANNs on simulated resistive crossbar arrays *arXiv preprint arXiv:1906.02698*

[16] Lee M C and Wong H Y 2013 Charge Loss Mechanisms of Nitride-Based Charge Trap Flash Memory Devices *IEEE Transactions on Electron Devices* **60** 3256-64

[17] Tsai W J, Zous N K, Liu C J, Liu C C, Chen C H, Tahui W, Pan S, Chih-Yuan L and Gu S H 2001 Data retention behavior of a SONOS type two-bit storage flash memory cell *2-5 Dec. 2001)* p 32.6.1-.6.4

[18] Janai M, Eitan B, Shappir A, Lusky E, Bloom I and Cohen G 2004 Data retention reliability model of NROM nonvolatile memory products *IEEE Transactions on Device and Materials Reliability* **4** 404-15

[19] Janai M, Shappir A, Bloom I and Eitan B 2008 Relaxation of localized charge in trapping-based nonvolatile memory devices *27 April-1 May 2008)* p 417-23

[20] Rastegari M, Ordonez V, Redmon J and Farhadi A 2016 Xnor-net: Imagenet classification using binary convolutional neural networks: Springer) p 525-42

[21] Daniels M W, Madhavan A, Talatchian P, Mizrahi A and Stiles M D 2019 Energy-efficient stochastic computing with superparamagnetic tunnel junctions *arXiv preprint arXiv:1911.11204*

[22] Hirtzlin T, Bocquet M, Penkovsky B, Klein J-O, Nowak E, Vianello E, Portal J-M and Querlioz D 2019 Digital Biologically Plausible Implementation of Binarized Neural Networks with Differential Hafnium Oxide Resistive Memory Arrays *arXiv preprint arXiv:1908.04066*